\documentclass[10pt,twocolumn,letterpaper]{article}

\usepackage[final]{cvpr}      %

\usepackage{graphicx}
\usepackage{amsmath}
\usepackage{amssymb}
\usepackage{booktabs}                                      %

\usepackage{algorithm}

\usepackage{threeparttable}
\usepackage{color}
\usepackage{colortbl}
\usepackage{multirow}
\usepackage[table,xcdraw]{xcolor}

\usepackage{comment}
\usepackage{bm}
\usepackage{etoolbox}                                      %
\usepackage{bbm}
\usepackage{amsthm}
\usepackage{amsfonts}                                      %
\usepackage{url}
\usepackage{nth}
\usepackage{cite}
\usepackage{balance}
\usepackage{courier}                                       %

\usepackage{listings}                                      %
\usepackage[]{algpseudocode}
\usepackage{upgreek}

\usepackage{wrapfig}

\usepackage[pagebackref,breaklinks,colorlinks,linkcolor=red,citecolor=citecolor,urlcolor=blue]{hyperref}
\definecolor{citecolor}{RGB}{34,139,34}
\definecolor{mydarkblue}{rgb}{0,0.08,1}
\definecolor{mydarkgreen}{rgb}{0.02,0.6,0.02}
\definecolor{mydarkred}{rgb}{0.8,0.02,0.02}
\definecolor{mydarkorange}{rgb}{0.40,0.2,0.02}
\definecolor{mypurple}{RGB}{111,0,255}
\definecolor{myred}{rgb}{1.0,0.0,0.0}
\definecolor{mygold}{rgb}{0.75,0.6,0.12}
\definecolor{myblue}{rgb}{0,0.2,0.8}
\definecolor{mydarkgray}{rgb}{0.,0.2,0.2}

\definecolor{lightred}{RGB}{255,235,235}
\definecolor{lightgreen}{RGB}{235,255,235}
\definecolor{lightblue}{RGB}{235,235,255}
\definecolor{lightcyan}{RGB}{235,255,255}
\definecolor{lightmagenta}{RGB}{255,235,255}
\definecolor{lightyellow}{RGB}{255,255,235}

\definecolor{qxkcolor}{RGB}{215,235,255}
\definecolor{softmaxcolor}{RGB}{230,235,255}
\definecolor{probxvcolor}{RGB}{255,255,235}

\definecolor{topkcolor}{RGB}{255,235,235}
\definecolor{zecolor}{RGB}{255,255,235}
\definecolor{dynacolor}{RGB}{235,255,255}

\definecolor{reviewcolor}{RGB}{0,0,200}

\newcommand{\ceil}[1]{\lceil #1 \rceil}

\newcommand{\calO}{\mathcal{O}}

\newcommand{\calP}{\mathcal{P}}

\theoremstyle{plain}

\theoremstyle{definition}

\algdef{SE}[DOWHILE]{Do}{doWhile}{\algorithmicdo}[1]{\algorithmicwhile\ #1}%

\newcommand{\squishlist}{
 \begin{list}{$\bullet$}
  { \setlength{\itemsep}{0pt}
     \setlength{\parsep}{3pt}
     \setlength{\topsep}{3pt}
     \setlength{\partopsep}{0pt}
     \setlength{\leftmargin}{1.5em}
     \setlength{\labelwidth}{1em}
     \setlength{\labelsep}{0.5em} } }
     
\newcommand{\squishend}{
  \end{list}  }

\newcommand{\ours}{\texttt{HRViT}\xspace}

\graphicspath{{./figs/}}

\def\cvprPaperID{4853} %
\def\confName{CVPR}
\def\confYear{2022}

\begin{document}

\title{
Multi-Scale High-Resolution Vision Transformer for Semantic Segmentation
}

\author
{
Jiaqi Gu$^{2}\thanks{Work done during an internship at Meta Reality Labs.}$,
Hyoukjun Kwon$^1$,
Dilin Wang$^1$,
Wei Ye$^1$,
Meng Li$^1$,
Yu-Hsin Chen$^1$,\\
Liangzhen Lai$^1$,
Vikas Chandra$^1$,
David Z. Pan$^2$\\
$^1$Meta Reality Labs, $^2$University of Texas at Austin\\

\small{\texttt{jqgu@utexas.edu}}
}

\maketitle
\begin{abstract}
\label{abstract}
Vision Transformers (ViTs) have emerged with superior performance on computer vision tasks compared to convolutional neural network (CNN)-based models.
However, ViTs are mainly designed for image classification that generate single-scale low-resolution representations, which makes dense prediction tasks such as semantic segmentation challenging for ViTs.
Therefore, we propose \ours, which enhances ViTs to learn semantically-rich and spatially-precise multi-scale representations by integrating high-resolution multi-branch architectures with ViTs.
We balance the model performance and efficiency of \ours by various branch-block co-optimization techniques.
Specifically, we explore heterogeneous branch designs, reduce the redundancy in linear layers, and augment the attention block with enhanced expressiveness.
Those approaches enabled \ours to push the Pareto frontier of performance and efficiency on semantic segmentation to a new level, as our evaluation results on ADE20K and Cityscapes show.
\ours achieves 50.20\% mIoU on ADE20K and 83.16\% mIoU on Cityscapes, surpassing state-of-the-art MiT and CSWin backbones with an average of \textbf{+1.78} mIoU improvement, \textbf{28\%} parameter saving, and \textbf{21\%} FLOPs reduction, demonstrating the potential of \ours as a strong vision backbone for semantic segmentation.
\end{abstract}

\section{Introduction}
\label{sec:Introduction}
Dense prediction tasks such as semantic segmentation are important computer vision workloads on emerging intelligent computing platforms, e.g., AR/VR devices.
Convolutional neural networks (CNNs) have rapidly evolved with significant performance improvement in semantic segmentation~\cite{NN_CVPR2015_Long, NN_ICLR2015_Chen, NN_ECCV2018_Chen, NN_MICCAI2015_Ronneberger, NN_TPAMI2021_Wang, NN_TPAMI2017_Badrinarayanan}.
Beyond classical CNNs, vision Transformers (ViTs) have emerged with competitive performance in computer vision tasks~\cite{NN_ECCV2020_Carion, NN_ICLR2021_Dosovitskiy,NN_Arxiv2021_Yuan, NN_Arxiv2021_Li,NN_ICCV2021_Zhang, NN_ICML2021_Touvron,NN_ICCV2021_Liu, NN_NeurIPS2021_Chu, NN_NeurIPS2021_Xie, NN_Arxiv2021_Dong, NN_Arxiv2021_Wang_CrossFormer, NN_CVPR2021_Wang,NN_ICCV2021_Chen,NN_ICCV2021_Xu}. 
Benefiting from the self-attention operations, ViTs embrace strong expressivity with long-distance information interaction and dynamic feature aggregation.
However, ViT~\cite{NN_ICLR2021_Dosovitskiy} produces single-scale and low-resolution representations, which are not friendly to semantic segmentation that requires high position sensitivity and fine-grained image details.

To cope with the challenge, various ViT backbones that yield multi-scale representations were proposed for semantic segmentation~\cite{NN_ICCV2021_Wang,NN_ICCV2021_Liu,NN_Arxiv2021_Yu,NN_Arxiv2021_Wang_CrossFormer,NN_NeurIPS2021_Chu,NN_NeurIPS2021_Xie,NN_Arxiv2021_Dong}.
However, they still follow a classification-like network topology with a \emph{sequential} or \emph{series} architecture.
Based on complexity consideration, they gradually downsample the feature maps to extract higher-level \emph{low-resolution (LR) representations} and directly feed each stage's output to the downstream segmentation head.
Such sequential structures lack enough cross-scale interaction thus cannot produce high-quality \emph{high-resolution (HR) representations}.

HRNet~\cite{NN_TPAMI2021_Wang} was proposed to solve the problem outside of ViT context, which enhances the cross-resolution interaction with a multi-branch architecture
maintaining all resolutions throughout the network.
HRNet extracts multi-resolution features in parallel and fuses them repeatedly to generate high-quality HR representations with rich semantic information. 
Such a design concept has achieved great success in various dense prediction tasks.
Nevertheless, its expressivity is limited by small receptive fields and strong inductive bias from cascaded convolution operations.
To deal with the challenge, some HRNet variants such as Lite-HRNet~\cite{NN_CVPR2021_Yu} and HR-NAS~\cite{NN_CVPR2021_Ding} are proposed.
However, those improved HRNet designs are still mainly based on the convolutional building blocks, and their demonstrated performance on semantic segmentation is still far behind the SoTA scores of ViT counterparts. 

Therefore, \emph{synergistically} integrating HRNet with ViTs is an approach to be explored for further performance improvement.
By combining those two approaches, ViTs can obtain rich multi-scale representability from the HR architecture, while HRNet can gain a larger receptive field from the attention operations.
However, migrating the success of HRNet to ViT backbones is non-trivial.
Given the high complexity of multi-branch HR architectures and self-attention operations, simply replacing all convolutional residual blocks in HRNet with Transformer blocks will encounter severe scalability issues.
The inherited high representation power from \emph{multi-scale} can be overwhelmed by the prohibitive latency and energy cost on hardware without careful architecture-block co-optimization.

Therefore, we propose \ours, an efficient \emph{multi-scale high-resolution} vision Transformer backbone specifically optimized for semantic segmentation.
\ours enables multi-scale representation learning in ViTs and improves the efficiency based on the following approaches:
(1) \ours's multi-branch HR architecture extracts multi-scale features in parallel with cross-resolution fusion to enhance the multi-scale representability of ViTs; 
(2) \ours's augmented local self-attention removes redundant keys and values for better efficiency and enhances the model expressivity with extra parallel convolution paths, additional nonlinearity units, and auxiliary shortcuts for feature diversity enhancement; 
(3) \ours adopts mixed-scale convolutional feedforward networks to fortify the multi-scale feature extraction; 
(4) \ours's HR convolutional stem and efficient patch embedding layers maintain more low-level fine-grained features with reduced hardware cost.
Also, distinguished from the HRNet-family, \ours follows a unique heterogeneous branch design to balance efficiency and performance, which is not simply an improved HRNet or a direct ensemble of HRNet and self-attention but a new topology of pure ViTs mainly constructed by self-attention with careful branch-block co-optimization.

Based on the approaches in \ours, we make the following contributions:
\squishlist
    {\item We deeply investigate the multi-scale representation learning in vision Transformers and propose \ours that integrates multi-branch high-resolution architectures with vision Transformers.}
    {\item To enhance the efficiency of \ours for scalable HR-ViT integration, we propose a set of approaches as follows: exploiting the redundancy in Transformer blocks, developing performance-efficiency co-optimized building blocks, and adopting heterogeneous branch designs.}
    {\item 
    We evaluate \ours on ADE20K and Cityscapes and present results that push the Pareto frontier of performance and efficiency forward as follows: }
    \ours achieves 50.20\% mIoU on ADE20K \texttt{val} and 83.16\% mIoU on Cityscapes \texttt{val} for semantic segmentation tasks, outperforming state-of-the-art (SoTA) MiT and CSWin backbones with 1.78 higher mIoU, 28\% fewer parameters, and 21\% lower FLOPs, on average.
\squishend

\section{Proposed HRViT Architecture}
\label{sec:Method}
\begin{figure*}
    \centering
    \includegraphics[width=0.99\textwidth]{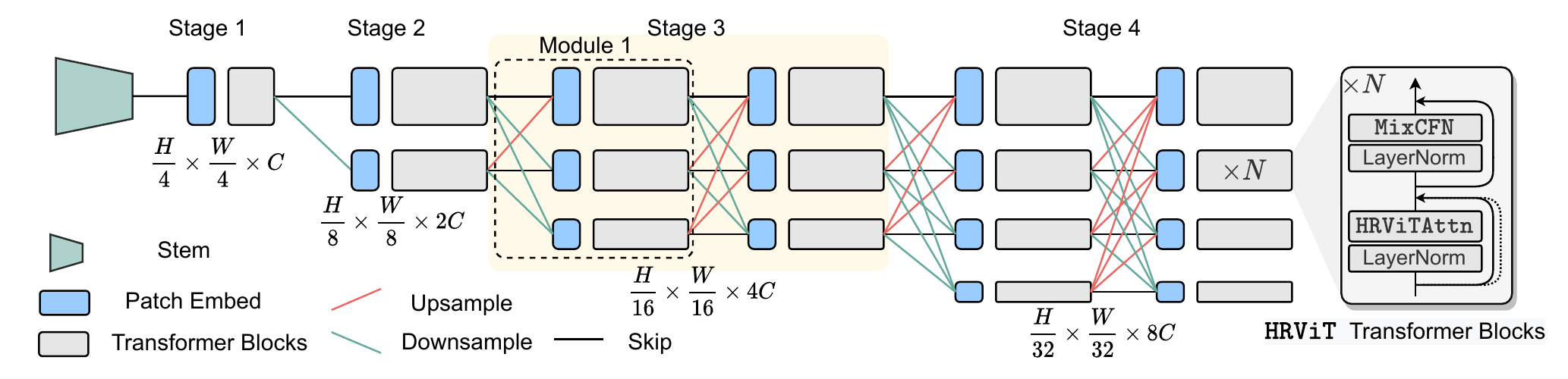}
    \caption{The overall architecture of our proposed \ours.
    It progressively expands to 4 branches.
    Each stage has multiple modules.
    Each module contains multiple Transformer blocks.
    }
    \label{fig:HRViTArch}
\end{figure*}

Recent advances in vision Transformer backbone designs mainly focus on attention operator innovations.
A new topology design can create another dimension to unleash the potential of ViTs with even stronger vision expressivity.
Extending the sequential topology of ViTs to the multi-branch structure, inspired by HRNet, is a promising approach for performance improvement.
An important question that remains to be answered is whether the \emph{success of HRNet can be efficiently migrated to ViT backbones} to consolidate their leading position in dense prediction tasks such as semantic segmentation. 

In this section, we delve into the multi-scale representation learning in ViTs and introduce an efficient integration of the HR architecture and Transformer.

\subsection{Architecture overview}
\label{sec:ArchOverview}
As illustrated in Figure~\ref{fig:HRViTArch}, the first part of \ours consists of a convolutional stem to reduce spatial dimensions while extracting low-level features.
After the convolutional stem, \ours deploys four progressive Transformer stages where the $n$-th stage contains $n$ \emph{parallel multi-scale} Transformer branches. 
Each stage can have one or more modules. 
Each module starts with a lightweight dense fusion layer to achieve cross-resolution interaction and an efficient patch embedding block for local feature extraction, followed by repeated augmented local self-attention blocks (\texttt{HRViTAttn}) and mixed-scale convolutional feedforward networks (\texttt{MixCFN}). 
Unlike sequential ViT backbones that progressively reduce the spatial dimension to generate pyramid features, we \emph{maintain the HR features throughout the network} to strengthen the quality of HR representations via cross-resolution fusion.

\subsection{Efficient HR-ViT integration with heterogeneous branch design}
\label{sec:OverallArchitecture}

We design a heterogeneous multi-branch architecture for efficient multi-scale high-resolution ViTs. 
A straightforward choice to fuse HRNet and ViTs is to replace all convolutions in HRNet with self-attentions.
However, given the high complexity of multi-branch HRNet and self-attention operators, this brute-force combining will quickly cause an explosion in memory footprint, parameter size, and computational cost.
Therefore, we need to carefully design the model architecture and building blocks to enable scalable and efficient HR-ViT integration.
We discuss our approaches to achieve the scalability and efficiency next.

\noindent\textbf{Heterogeneous branch configuration.}~
For the branch architecture in \ours, we need to determine the number of Transformer blocks assigned for each branch.
Simply assigning the same number of blocks with the same local self-attention window size on each module will result in intractably large computational costs.
Therefore, we analyze the functionality and cost of each branch in Table~\ref{tab:BranchAnalysis}, and we propose a simple design heuristic based on the analysis.

\begin{table}[]
\centering
\resizebox{.49\textwidth}{!}{
\begin{tabular}{c|ccc}
\toprule
Feature/Arch.            & HR ($\frac{1}{4}\times,\frac{1}{8}\times$)                       & MR ($\frac{1}{16}\times$)           & LR ($\frac{1}{32}\times$)                    \\ \midrule
Memory cost                     & High                     & Medium       & Low                   \\
Computation                     & Heavy                    & Moderate     & Light                 \\
\#Params                        & Small                    & Medium       & Large                 \\
Eff. on class. & Not quite useful         & Important    & Important             \\
Feat. granularity     & Fine       & Medium       & Coarse \\
Receptive field  & Local   & Region  &  Global\\
\midrule
Window size                     & Narrow ($s$=1,2)           & Wide ($s$=7)   & Wide ($s$=7)            \\
Depth                  & Shallow ($\sim$5-6) & Deep (20-30) & Shallow ($\sim$4)         \\ \bottomrule
\end{tabular}
}
\caption{
Qualitative cost and functionality analysis.
Window sizes and depth are given for each branch.
\emph{Eff. on class.} and \emph{Feat. granularity} are short for effectiveness on image-level classification and feature granularity.
}
\label{tab:BranchAnalysis}
\end{table}

We analyze (1) the number of parameters and (2) the number of floating-point operations (FLOPs) in \texttt{HRViTAttn} and \texttt{MixCFN} blocks on the $i$-th branch ($i=1,2,3,4$) as follows:
\begin{equation}
    \small
    \label{eq:ParamFLOPComplexity}
    \begin{aligned}
    \texttt{Params}_{\texttt{HRViTAttn},i}&=\calO(4^{i-1}C^2+2^{i-1}C),\\
    \texttt{Params}_{\texttt{MixCFN},i}&=\calO(4^{i-1}C^2r_i+2^{i-1}Cr_i),\\
    \texttt{FLOPs}_{\texttt{HRViTAttn},i}&=\calO\Big(HWC^2\!+\!\frac{CHW(H\!+\!W)s_i}{4^{i-1}}\Big),\\
    \texttt{FLOPs}_{\texttt{MixCFN},i}&=\calO\Big(r_iHWC^2+\frac{r_iHWC}{2^{i-1}}\Big).
    \end{aligned}
\end{equation}

We use Equation \ref{eq:ParamFLOPComplexity} to compare the memory cost, the computation, the number of parameters, and computation in Table~\ref{tab:BranchAnalysis}.

Based on the complexity analysis, we observe that the first and second HR branches ($i\!=\!1,2$) involve a high memory and computational cost. 
Hence, those HR branches typically can not afford a large enough receptive field for image-level classification.
On the other hand, they are parameter-efficient and able to provide fine-grained detail calibration in segmentation tasks. 
Thus, we \emph{use a narrow attention window size and use a minimum number of blocks on two HR paths}.

We observe that the most important branch is the third one with a medium resolution (MR). 
Given its medium hardware cost, we can afford \emph{a deep branch with a large window size on the MR path} to provide large receptive fields and well-extracted high-level features.

The lowest resolution (LR) branch contains most parameters and is very useful to provide high-level features with a global receptive field to generate coarse segmentation maps. 
However, its small spatial sizes result in too much loss of image details.
Therefore, we only deploy \emph{a few blocks with a large window size on the LR branch} to improve high-level feature quality under parameter budgets.

\noindent\textbf{Nearly-even block assignment.}~
A unique problem in \ours is to determine \emph{how to assign blocks to each module}
In \ours, we need to assign 20 blocks to 4 modules on the 3rd path. 
To maximize the average depth of the network ensemble and help input/gradient flow through the deep Transformer branch, we employ a \emph{nearly-even partitioning}, e.g., 6-6-6-2, and exclude an extremely unbalanced assignment, e.g., 17-1-1-1.

\subsection{Efficient HRViT component design}
We discussed how we enable HR-ViT integration with heterogeneous branch design.
Next, we discuss how to further push the efficiency and performance boundary via block optimization.

\begin{figure}
    \centering
    \subfloat[]{\includegraphics[width=0.48\textwidth]{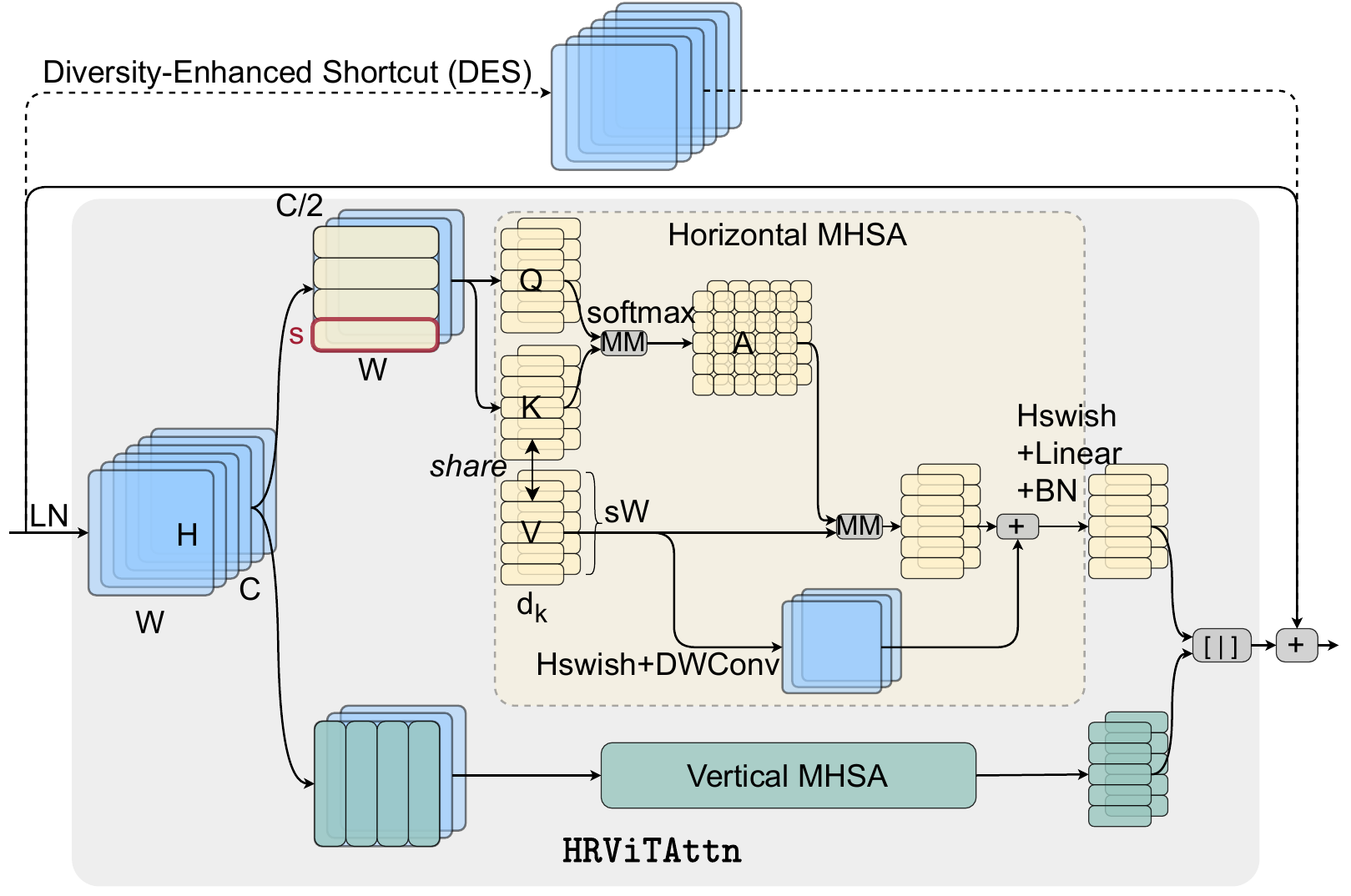}
    \label{fig:HRViTAttentionOp}
    }\\
    \subfloat[]{\includegraphics[width=0.45\textwidth]{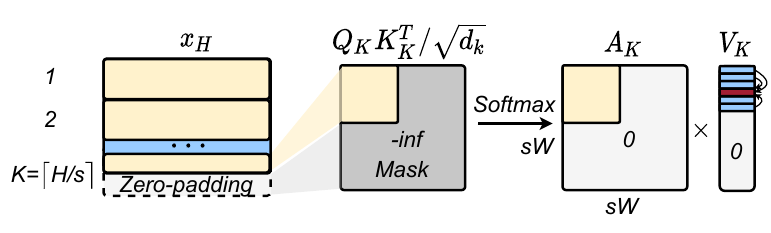}
    \label{fig:AttentionMask}
    }
    \caption{(a) \texttt{HRViTAttn}: augmented cross-shaped local self-attention with a parallel convolution path and an efficient diversity-enhanced shortcut.
    (b) Window zero-padding with attention map masking.}
    \label{fig:HRViTAttention}
\end{figure}

\noindent\textbf{Augmented cross-shaped local self-attention.}
To achieve high performance with improved efficiency, a hardware-efficient self-attention operator is necessary.
We adopt one of the SoTA efficient attention designs, cross-shaped self-attention~\cite{NN_Arxiv2021_Dong},
as our baseline attention operator.
Based on that, we design our \emph{augmented cross-shaped local self-attention} \texttt{HRViTAttn} illustrated in Figure~\ref{fig:HRViTAttention}, which provides the following benefits:
(1) \emph{Fine-grained attention}: 
Compared to globally-downsampled attentions~\cite{NN_ICCV2021_Wang,NN_NeurIPS2021_Xie}, this one has fine-grained feature aggregation that preserves detailed information.
(2) \emph{Approximate global view}: 
By using two parallel orthogonal local attentions, this attention can collect global information.
(3) \emph{Scalable complexity}: 
one dimension of the window is fixed, which avoids quadratic complexity to image sizes.

To balance the performance and efficiency, we introduce our augmented version, denoted as \texttt{HRViTAttn}, with several key optimizations.

In Figure~\ref{fig:HRViTAttentionOp}, we follow the cross-shaped window partitioning approach in CSWin that separates the input $x\in\mathbb{R}^{H\times W\times C}$ into two parts $\{x_H,x_V\in\mathbb{R}^{H\times W\times C/2}\}$. 
$x_H$ is partitioned into disjoint horizontal windows, and the other half $x_V$ is chunked into vertical windows. 
The window is set to $s\times W$ or $H \times s$.
Within each window, the patch is chunked into $K$ $d_k$-dimensional heads, then a local self-attention is applied,
\begin{equation}
\small
\label{eq:HRViTAttention}
\begin{aligned}
\texttt{HRViTAttn}(x)&=\texttt{BN}\big(\sigma(W^O[y_1,\cdots,y_k,\cdots,y_K])\big)\\
y_k&=z_k+\texttt{DWConv}\big(\sigma(W_k^V x)\big)\\
[z_k^1,\cdots,z_k^M]=z_k&=\left\{
\begin{aligned}
\texttt{H-Attn}_k(x), \quad 1\leq k<K/2\\
\texttt{V-Attn}_k(x), \quad K/2\leq k\leq K
\end{aligned}
\right.\\
z_k^m &=\texttt{MHSA}(W_k^Q x^m,W_k^K x^m,W_k^V x^m)\\
[x^1,\cdots,x^m,\cdots,x^M]&=x, \quad x^m\in\mathbb{R}^{(H/s) \times W\times C},
\end{aligned}
\end{equation}
where $W_k^Q,W_k^K,W_k^V\in\mathbb{R}^{d_k\times C}$ are projection matrices to generate query $Q_k$, key $K_k$, and value $V_k$ tensors for the $k$-th head, $W^O\in\mathbb{R}^{C\times C}$ is the output projection matrix, and $\sigma$ is Hardswish activation.
If the image sizes are not a multiple of window size, e.g., $s\ceil{H/s}>H$, we apply zero-padding to inputs $x_H$ or $x_V$ to allow a complete $K$-th window, shown in Figure~\ref{fig:AttentionMask}. 
Then the padded region in the attention map is masked to 0 to avoid incoherent semantic correlation. 

The original QKV linear layers are quite costly in computation and parameters. 
We share the linear projections for key and value tensors in \texttt{HRViTAttn} to save computation and parameters as follows, 
\begin{equation}
    \small
    \label{eq:ShareKV}
    \texttt{MHSA}(W^Q_k\!x^m,W^V_k\!x^m, W^V_k\!x^m)\!=\!\texttt{softmax}\Big(\frac{Q_k^m~(V_k^m)^T}{\sqrt{d_k}}\Big)V_k^m,
\end{equation}

In addition, we introduce an auxiliary path with parallel depth-wise convolution to inject inductive bias to facilitate training. 
Unlike the local positional encoding in CSWin, our parallel path is \emph{nonlinear and applied on the entire 4-D feature map $W^Vx$ without window-partitioning}.
This path can be treated as an inverted residual module sharing point-wise convolutions with the linear projection layers in self-attention. 
This shared path can effectively inject inductive bias and reinforce local feature aggregation with marginal hardware overhead.

As a performance compensation for the above key-value sharing, we introduce an extra Hardswish function to improve the nonlinearity. 
We also append a BatchNorm (BN) layer that is initialized to an identity projection to stabilize the distribution for better trainability.
Motivated by recent studies on the importance of shortcuts in ViTs~\cite{NN_Arxiv2021_Raghu}, we add a channel-wise projector as a diversity-enhanced shortcut (DES).
Unlike the augmented shortcut~\cite{NN_NeurIPS2021_Tang},
our shortcut has higher nonlinearity and does not depend on hardware-unfriendly Fourier transforms.
The projection matrix in our DES $\calP^{C\times C}$ is approximated by Kronecker decomposition $\calP=A^{\sqrt{C}\times\sqrt{C}}\otimes B^{\sqrt{C}\times\sqrt{C}}$ to minimize parameter cost.
Then we fold $x$ as $\tilde{x}\in\mathbb{R}^{HW\times\sqrt{C}\times\sqrt{C}}$ and convert $(A\otimes B) x$ into $(A\tilde{x}B^T)$ to save computations.
We further insert \texttt{Hardswish} after the $B$ projection to increase the nonlinearity,
\begin{equation}
    \label{eq:Shortcut}
    \texttt{DES}(x)=A\cdot\texttt{Hardswish}(\tilde{x}B^T).
\end{equation}

\begin{figure}
    \centering
    \includegraphics[width=0.48\textwidth]{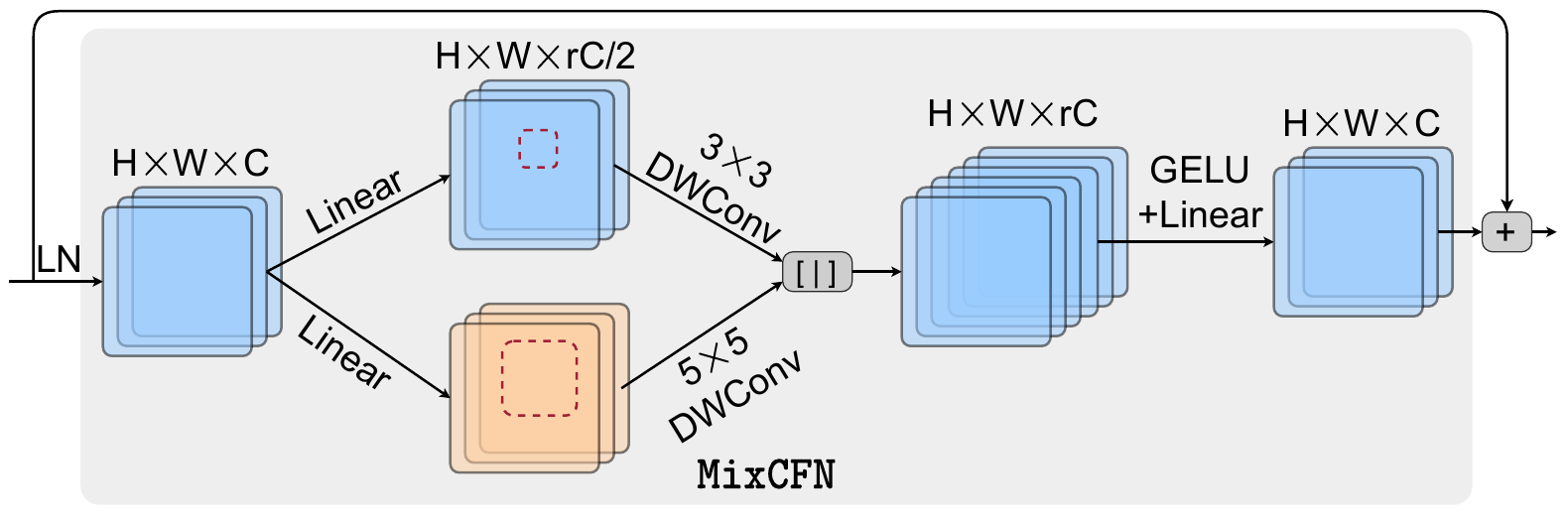}
    \caption{\texttt{MixCFN} with multiple depth-wise convolution paths to extract multi-scale local information.}
    \label{fig:MixCFN}
\end{figure}

\noindent\textbf{Mixed-scale convolutional feedforward network.}~
Inspired by the \emph{MixFFN} in MiT~\cite{NN_NeurIPS2021_Xie} and multi-branch inverted residual blocks in HR-NAS~\cite{NN_CVPR2021_Ding}, we design a mixed-scale convolutional FFN (\texttt{MixCFN}) by inserting two multi-scale depth-wise convolution paths between two linear layers, shown in Figure~\ref{fig:MixCFN}.

After LayerNorm, we expand the channel by a ratio of $r$, then split it into two branches. 
The 3$\times$3 and 5$\times$5 depth-wise convolutions (DWConvs) are used to increase the multi-scale local information extraction of \ours.
For efficiency consideration, we exploit the channel redundancy by reducing the \texttt{MixCFN} expansion ratio $r$ from 4~\cite{NN_NeurIPS2021_Xie,NN_ICCV2021_Liu} to 2 or 3 with marginal performance loss on medium to large models.

\noindent\textbf{Downsampling stem.}~
In semantic segmentation tasks, images are of high resolution, e.g., 1024$\times$1024.
Self-attention operators are known to be expensive as their complexity is quadratic to image sizes.
To address the scalability issue when processing large images, we down-sample the inputs by 4$\times$ before feeding into the main body of \ours.
We do not use attention operations in the stem since early convolutions are more effective to extract low-level features than self-attentions~\cite{NN_ICCV2021_Graham, NN_NeurIPS2021_Xiao}.
As early convolutions, we follow the design in HRNet and use two stride-2 CONV-BN-ReLU blocks as a stronger downsampling stem to extract $C$-channel features with more information maintained, unlike prior ViTs~\cite{NN_NeurIPS2021_Xie, NN_NeurIPS2021_Chu, NN_ICCV2021_Liu} that used a stride-4 convolution.

\noindent\textbf{Efficient patch embedding.}~
Before Transformer blocks in each module, we 
add a patch embedding block (CONV-LayerNorm) on each branch, which is used to match channels and extract patch information with enhanced inter-patch communication. 
However, the patch embedding layers have a non-trivial hardware cost in the HR architecture since each module at stage-$n$ will have $n$ embedding blocks. 
Therefore, we simplify the patch embedding to be a point-wise CONV followed by a depth-wise CONV~\cite{NN_CVPR2020_Haase},
\begin{equation}
    \small
    \label{eq:PatchEmbed}
    \texttt{EffPatchEmbed}(x)=\texttt{LN}\big(\texttt{DWConv}(\texttt{PWConv}(x))\big).
\end{equation}

\begin{figure}
    \centering
    \includegraphics[width=0.48\textwidth]{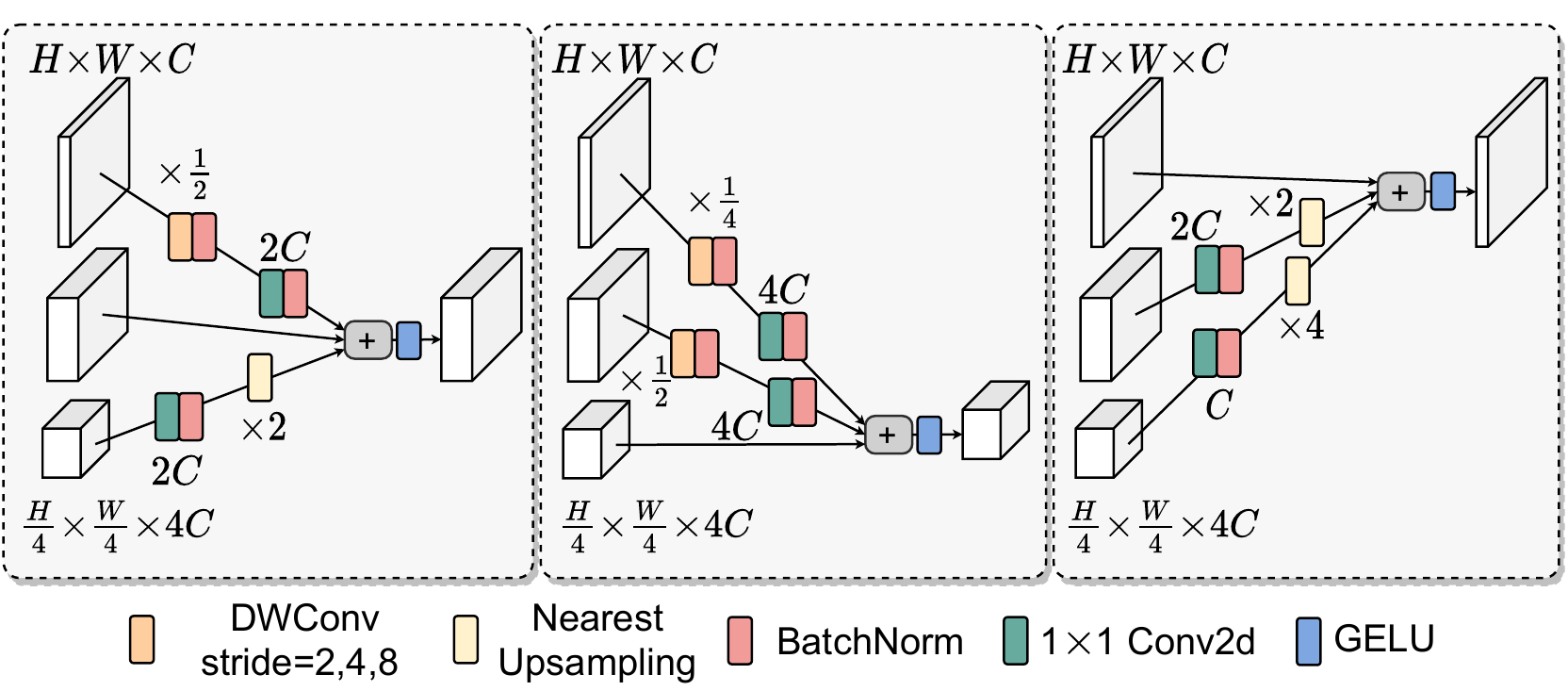}
    \caption{Cross-resolution fusion layers with channel matching, up-scaling, and down-sampling.}
    \label{fig:FusionLayer}
\end{figure}

\noindent\textbf{Cross-resolution fusion layer.}~
The cross-resolution fusion layer is critical for \ours to learn high-quality HR representations, shown in Figure~\ref{fig:FusionLayer}.
To enhance cross-resolution interaction, we insert repeated cross-resolution fusion layers at the beginning of each module following the approach in HRNet~\cite{NN_TPAMI2021_Wang, NN_CVPR2021_Yu}.

To \emph{help LR features maintain more image details and precise position information}, we merge them with down-sampled HR features. 
Instead of using a progressive convolution-based downsampling path to match tensor shapes~\cite{NN_TPAMI2021_Wang, NN_CVPR2021_Yu}, we employ a direct down-sampling path to minimize hardware overhead. 
In the down-sampling path between the $i$-th input and $j$-th output ($j>i$), we use a depth-wise separable convolution with a stride of $2^{j-i}$ to shrink the spatial dimension and match the output channels.
The kernel size used in the DWConv is ($2^{j-i}$+1) to create patch overlaps.
Those HR paths inject more image information into the LR path to \emph{mitigate information loss} and \emph{fortify gradient flows} during backpropagation to facilitate the training of deep LR branches.

On the other hand, the receptive field is usually limited in the HR blocks as we minimize the window size and branch depth on HR paths.
Hence, we merge LR representations into HR paths to help them \emph{obtain higher-level features with a larger receptive field}.
Specifically, in the up-scaling path ($j<i$), we first increase the number of channels with a point-wise convolution and up-scale the spatial dimension via a nearest neighbor interpolation with a rate of $2^{i-j}$.
When $i$=$j$, we directly pass the features to the output as a skip connection.
Note that in HR-NAS~\cite{NN_CVPR2021_Ding}, the dense fusion is simplified by a sparse fusion module where only neighboring resolutions are merged.
This technique is not considered in \ours since it saves marginal hardware cost but leads to a noticeable accuracy drop, which will be discussed in subsection~\ref{sec:Ablation}.

\subsection{Architectural variants}
\label{sec:ArchVariants}

\begin{table*}[]
\centering
\resizebox{0.99\textwidth}{!}{
\begin{tabular}{c|ccccc}
\toprule
\multicolumn{1}{l|}{Variant} & Architecture design          & Window $s$                                           & MixCFN ratio $r$                                           & \multicolumn{1}{l}{Channel $C$}                                & Head dim $d_k$                                                  \\ \midrule
\ours-b1              &  \begin{minipage}{.40\textwidth}
      \includegraphics[width=1.0\linewidth, height=19mm]{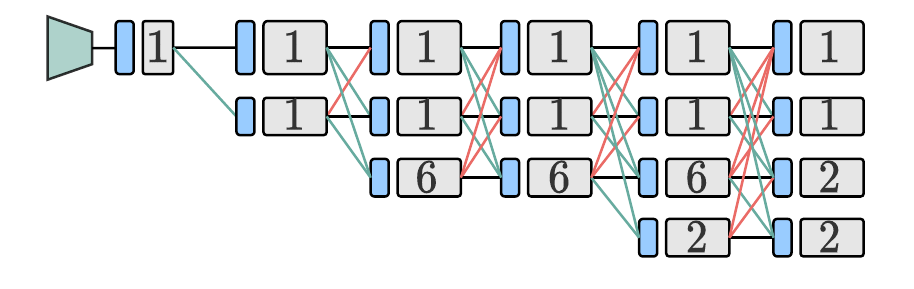}
    \end{minipage} & \begin{tabular}[c]{@{}c@{}}1\\ 2\\ 7\\ 7\end{tabular} & \begin{tabular}[c]{@{}c@{}}4\\ 4\\ 4\\ 4\end{tabular} & \begin{tabular}[c]{@{}c@{}}32\\ 64\\ 128\\ 256\end{tabular}  & \begin{tabular}[c]{@{}c@{}}16\\ 32\\ 32\\ 32\end{tabular} \\\midrule
\ours-b2              &  \begin{minipage}{.40\textwidth}
      \includegraphics[width=1.0\linewidth, height=19mm]{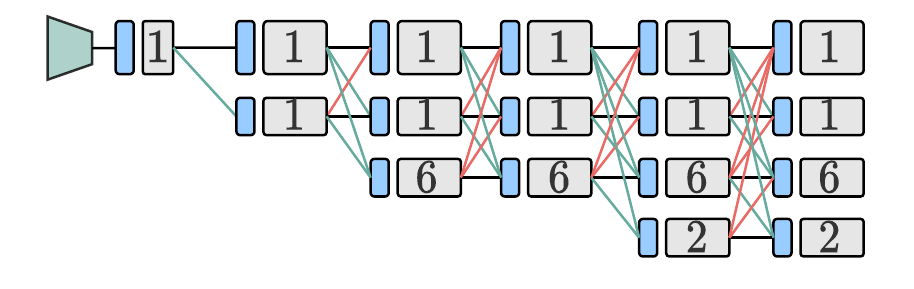}
    \end{minipage}  & \begin{tabular}[c]{@{}c@{}}1\\ 2\\ 7\\ 7\end{tabular} & \begin{tabular}[c]{@{}c@{}}2\\ 3\\ 3\\ 3\end{tabular} & \begin{tabular}[c]{@{}c@{}}48\\ 96\\ 240\\ 384\end{tabular}  & \begin{tabular}[c]{@{}c@{}}24\\ 24\\ 24\\ 24\end{tabular} \\\midrule
\ours-b3              & \begin{minipage}{.40\textwidth}
      \includegraphics[width=1.0\linewidth, height=19mm]{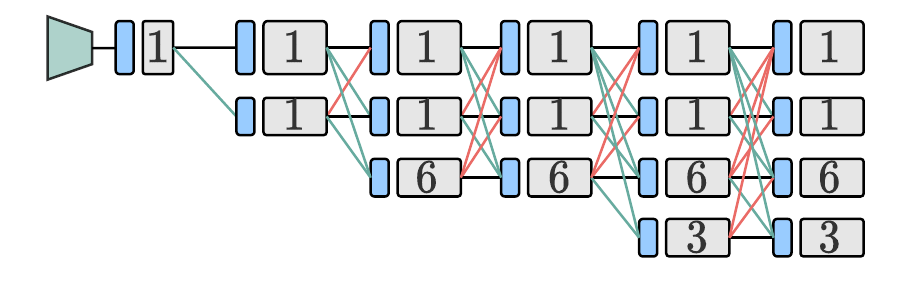}
    \end{minipage}  & \begin{tabular}[c]{@{}c@{}}1\\ 2\\ 7\\ 7\end{tabular} & \begin{tabular}[c]{@{}c@{}}2\\ 2\\ 2\\ 2\end{tabular} & \begin{tabular}[c]{@{}c@{}}64\\ 128\\ 256\\ 512\end{tabular} & \begin{tabular}[c]{@{}c@{}}32\\ 32\\ 32\\ 32\end{tabular} \\
    \bottomrule
\end{tabular}
}
\caption{Architecture variants of \ours.
The number of Transformer blocks is marked in each module, followed by
per branch settings.
}
\label{tab:ArchVariants}
\end{table*}

As shown in Table~\ref{tab:ArchVariants} with three design variants of \ours, variants of \ours scale in both network depth and width.
We follow the aforementioned design guidance and evenly assign 5-6 Transformer blocks to HR branches, 20-24 blocks to the MR branch, and 4-6 blocks to the LR branch.
Window sizes are set to (1,2,7,7) for 4 branches.
We use relatively large \texttt{MixCFN} expansion ratios in small variants for higher performance and reduce the ratio to 2 on larger variants for better efficiency.
We gradually follow the scaling rule from CSWin~\cite{NN_Arxiv2021_Dong} to increase the basic channel $C$ for the highest resolution branch from 32 to 64.
\#Blocks and \#channels can be flexibly tuned for the \emph{3rd/4th branch} to match a specific hardware cost.

\section{Experiments}
\label{sec:ExperimentalResults}
We pretrain all models on ImageNet-1K~\cite{NN_ImageNet1k} and conduct experiments on ADE20K~\cite{NN_ADE20K} and Cityscapes~\cite{NN_Cityscapes} for semantic segmentation.
We compare the performance and efficiency of our \ours with SoTA ViT backbones, i.e., Swin~\cite{NN_ICCV2021_Liu}, Twins~\cite{NN_NeurIPS2021_Chu}, MiT~\cite{NN_NeurIPS2021_Xie}, and CSWin~\cite{NN_Arxiv2021_Dong}.

\subsection{Semantic segmentation on ADE20K and Cityscapes}
\label{sec:SegmentationExp}
\begin{table}[]
\resizebox{0.48\textwidth}{!}{
\begin{tabular}{l|cccc}
\toprule
\multirow{2}{*}{Variant}    & \multirow{2}{*}{Image Size}  & \multirow{2}{*}{\begin{tabular}[c]{@{}c@{}}\#Params \\ (M)\end{tabular}}  & \multirow{2}{*}{GFLOPs} & \multirow{2}{*}{\begin{tabular}[c]{@{}c@{}}IMNet-1K \\ top-1 acc.\end{tabular}} \\\\ \midrule
\ours-b1    & 224        & 19.7         & 2.7    & 80.5                                                           \\
\ours-b2    & 224        & 32.5         & 5.1    & 82.3                                                           \\
\ours-b3    & 224        & 37.9         & 5.7    & 82.8                                                           \\
\bottomrule
\end{tabular}
}
\caption{ImageNet-1K pre-training results of \ours.
FLOPs are measured on an image size of 224$\times$224.
\#Params includes the classification head as used in HRNetV2~\cite{NN_TPAMI2021_Wang}.
}
\label{tab:CompareImageNet1K}
\end{table}

On semantic segmentation, \ours achieves the best performance-efficiency Pareto front, surpassing the SoTA MiT and CSWin backbones. 
\ours (b1-b3) outperform the previous SoTA SegFormer-MiT (B1-B3)~\cite{NN_NeurIPS2021_Xie} with \textbf{+3.68}, \textbf{+2.26}, and \textbf{+0.80} higher mIoU on ADE20K \texttt{val}, and \textbf{+3.13}, \textbf{+1.81}, \textbf{+1.46} higher mIoU on Cityscapes \texttt{val}.

\noindent\textbf{ImageNet-1K pre-training.}~
All \ours variants are pre-trained on ImageNet-1K, shown in Table~\ref{tab:CompareImageNet1K}.
We follow the same pre-training settings as DeiT~\cite{NN_ICML2021_Touvron} and other ViTs~\cite{NN_ICCV2021_Liu,NN_NeurIPS2021_Xie, NN_Arxiv2021_Dong}.
We adopt stochastic depth~\cite{NN_ECCV2016_Huang} for all \ours variants with the max drop rate of 0.1.
The drop rate is gradually increased on the deepest 3rd branch, and other shallow branches follow the rate of the 3rd branch within the same module.
We use the HRNetV2~\cite{NN_TPAMI2021_Wang} classification head in \ours on ImageNet-1K pre-training.
More details can be found in Appendix~\ref{sec:AppendixImageNet}.

\noindent\textbf{Settings.}~
We evaluate \ours for semantic segmentation on the Cityscapes and ADE20K datasets. 
We employ a lightweight SegFormer~\cite{NN_NeurIPS2021_Xie} head 
based on the mmsegmentation framework~\cite{NN_mmseg}. 
We follow the training settings of prior work~\cite{NN_NeurIPS2021_Xie, NN_Arxiv2021_Dong}.
The training image size for ADE20K and Cityscapes are 512$\times$512 and 1024$\times$1024, respectively.
The test image size for ADE20K and Cityscapes is set to 512$\times$2048 and 1024$\times$2048, respectively.
We do inference on Cityscapes with sliding window test by cropping 1024$\times$1024 patches.
More details are in Appendix~\ref{sec:AppendixSegmentation}.

\begin{figure*}
    \centering
    \includegraphics[width=0.98\textwidth]{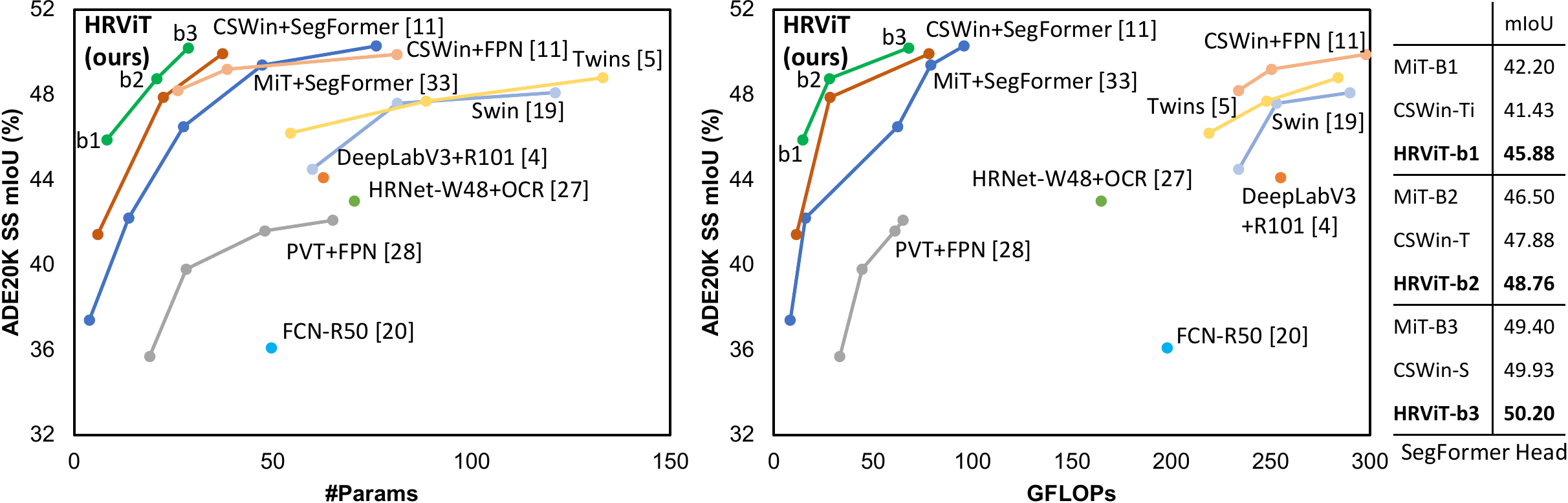}
    \caption{\ours achieves the best performance-efficiency trade-off among all models on ADE20K \texttt{val}.
    The table on the right shows ADE20K \texttt{val} mIoUs of MiT, CSWin, and \ours with the SegFormer~\cite{NN_NeurIPS2021_Xie} head.}
    \label{fig:ADE20KCurveTable}
    \vspace{-10pt}
\end{figure*}

\noindent\textbf{Results on ADE20K.}~
We evaluate different ViT backbones in single-scale mean intersection-over-union (mIoU), \#Params, and GFLOPs. 
Figure~\ref{fig:ADE20KCurveTable} plots the Pareto curves in the \#Params and FLOPs space.
On ADE20K \texttt{val}, \ours outperforms other ViTs with better performance and efficiency trade-off.
For example, with the SegFormer head, \ours-b1 outperforms MiT-B1 with 3.68\% higher mIoU, 40\% fewer parameters, and 8\% less computation.
Our \ours-b3 achieves a higher mIoU than the best CSWin-S but saves 23\% parameters and 13\% FLOPs. 
Compared with HRNetV2+OCR, our \ours shows considerable performance and efficiency advantages. 
We also evaluate \ours with UperNet~\cite{NN_ECCV2018_Xiao} head in Appendix~\ref{sec:AppendixUperNet}.

\noindent\textbf{Results on Cityscapes.}~
We summarize the results on Cityscapes in Table~\ref{tab:CompareCityscapes}.
Our small model \ours-b1 outperforms MiT-B1 and CSWin-Ti by +3.13 and +2.47 higher mIoU.
The key insight is that the HR architecture can increase the effective width of models with narrow channels, leading to higher modeling capacity.
Hence, the parallel multi-branch topology is especially beneficial for small networks.
When training \ours-b3 on Cityscapes, we set the window sizes to 1-2-9-9.
\ours-b3 outperforms the MiT-b4 with +0.86 higher mIoU, 55.4\% fewer parameters, and 30.7\% fewer FLOPs.
Compared with two SoTA ViT backbones, i.e., MiT and CSWin, \ours achieves an average of +2.16 higher mIoU with 30.7\% fewer parameters and 23.1\% less computation.

\begin{table}[]
\centering
\resizebox{0.48\textwidth}{!}{
\begin{tabular}{l|ccc}
\toprule
                           & \multicolumn{3}{c}{SegFormer Head~\cite{NN_NeurIPS2021_Xie}} \\
\multirow{-2}{*}{Backbone} & \#Param. (M)$\downarrow$  & GFLOPs$\downarrow$  & mIoU (\%)$\uparrow$ \\ \midrule
MiT-B0~\cite{NN_NeurIPS2021_Xie}                     & 3.8           & 8.4     & 76.20     \\
MiT-B1~\cite{NN_NeurIPS2021_Xie}                     & 13.7          & 15.9    & 78.50     \\
CSWin-Ti~\cite{NN_Arxiv2021_Dong}                   & 5.9           & 11.4    & 79.16     \\
\textbf{\ours-b1}                   & \textbf{8.1}           & \textbf{14.1}    & \textbf{81.63}     \\ \midrule
MiT-B2~\cite{NN_NeurIPS2021_Xie}                     & 27.5          & 62.4    & 81.00     \\
CSWin-T~\cite{NN_Arxiv2021_Dong}                    & 22.4          & 28.3    & 81.56     \\
\textbf{\ours-b2}                   & \textbf{20.8}          & \textbf{27.4}    & \textbf{82.81}     \\ \midrule
MiT-B3~\cite{NN_NeurIPS2021_Xie}                     & 47.3          & 79.0    & 81.70     \\
MiT-B4~\cite{NN_NeurIPS2021_Xie}                     & 64.1          & 95.7    & 82.30     \\
CSWin-S~\cite{NN_Arxiv2021_Dong}                    & 37.3          & 78.1    & 82.58     \\
\textbf{\ours-b3}                   & \textbf{28.6}          & \textbf{66.8}    & \textbf{83.16}     \\
\bottomrule

Avg improv.                   & -30.7\%          & -23.1\%    & +2.16     \\
\bottomrule
\end{tabular}
}
\caption{Comparison on the Cityscapes \texttt{val} segmentation dataset.
We reduce the channels (64$\rightarrow$32) of CSWin-T and name it CSWin-Ti.
FLOPs are based on the image size of 512$\times$512.}
\label{tab:CompareCityscapes}
\end{table}

\subsection{Ablation studies}
\label{sec:Ablation}
In Table~\ref{tab:Ablation}, we first compare with a baseline where all block optimization techniques are removed.
Our proposed key techniques can \emph{synergistically} improve the ImageNet accuracy by 0.73\% and the Cityscapes mIoU by +1.18 with 20\% fewer parameters and 13\% less computation.
Then we \emph{independently remove each technique} from \ours to validate their individual contribution.

\noindent\textbf{Sharing key-value.}~
When removing key-value sharing, i.e., using independent keys and values, \ours-b1 shows the same ImageNet-1K accuracy but at the cost of lower Cityscapes segmentation mIoU, 9\% more parameters, and 4\% more computations.
\begin{table}[]
\resizebox{0.48\textwidth}{!}{
\begin{tabular}{l|cccc}
\toprule
\multirow{2}{*}{Variants} & \multirow{2}{*}{\begin{tabular}[c]{@{}c@{}}\#Params \\ (M)\end{tabular}} & \multirow{2}{*}{\begin{tabular}[c]{@{}c@{}}FLOPs\\ (G)\end{tabular}} & \multirow{2}{*}{\begin{tabular}[c]{@{}c@{}}IMNet\\ top-1 acc.\end{tabular}} & \multirow{2}{*}{\begin{tabular}[c]{@{}c@{}}City\\ mIoU\end{tabular}}\\\\ \midrule
\textbf{\ours-b1}                                     & \textbf{8.1}         & \textbf{14.1}    & \textbf{80.52}    &  \textbf{81.63}                                                \\ \midrule
$-$ Key-value sharing                               & 8.8         & 14.7    & 80.52 & 81.00                                                     \\
$-$ Eff. patch embed & 9.9         & 16.5    & 80.19                                 &  81.18                   \\
$-$ MixCFN                                   & 7.9         & 13.6    & 79.86          & 80.52                                           \\
$-$ Parallel CONV path                       & 8.1         & 14.0    & 80.06           &  80.82                                         \\
$-$ Nonlinearity/BN                          & 8.1         & 14.1    & 80.37           &   81.12                                        \\
$-$ Dense fusion                             & 8.0 & 14.0 &  79.95 & 81.26\\
$-$ DES                             & 8.1 & 14.0 & 80.36  & 81.38\\\bottomrule
$-$ All block opt.                             & 10.1 & 16.3 & 79.79  & 80.45\\\bottomrule
\end{tabular}
}
\caption{Ablation on proposed techniques.
Each entry removes one technique independently.
The last one removes all techniques.
}
\label{tab:Ablation}
\vspace{-5pt}
\end{table}

\noindent\textbf{Patch embedding.}~
Changing our \texttt{EffPathEmbed} to the CONV-based counterpart~\cite{NN_NeurIPS2021_Xie} leads to 22\% more parameters and 17\% more FLOPs without accuracy/mIoU benefits.

\noindent\textbf{MixCFN.}~
Replacing the \texttt{MixCFN} block with the original FFNs~\cite{NN_ICLR2021_Dosovitskiy} directly leads to $\sim$0.66\% ImageNet accuracy drop and 0.11 Cityscapes mIoU loss with marginal efficiency improvement.
By adding multi-scale local feature extraction in feedforward networks, \texttt{MixCFN} can indeed boost the performance of \ours.

\noindent\textbf{Parallel convolution path.}~
The embedded inverted residual path in the \texttt{HRViTAttn} block is very lightweight and contributes 0.46\% higher ImageNet accuracy as well as 0.81 higher mIoU on Cityscapes.

\noindent\textbf{Additional nonlinearity/BN.}~
The extra Hardswish and BN introduce negligible overhead but boost expressivity and trainability, bringing 0.15\% higher ImageNet-1K accuracy 0.51 higher mIoU on Cityscapes \texttt{val}.

\noindent\textbf{Dense vs. sparse fusion layers.}~
The sparse fusion layer proposed in HR-NAS~\cite{NN_CVPR2021_Ding} is not very effective in \ours as it saves tiny hardware cost ($<$1\%) but leads to 0.57\% accuracy drop and 0.37 mIoU loss.

\noindent\textbf{Diversity-enhanced shortcut.}~
As an auxiliary path, the proposed shortcut (DES) helps enhance the feature diversity and effectively boosts the performance to a higher level both on classification and segmentation tasks.
The hardware overhead is negligible due to the high efficiency of the Kronecker decomposition-based projector.

\noindent\textbf{Vanilla HRNet-ViT baselines vs. \ours.}~
\begin{table}[]
\centering
\resizebox{0.48\textwidth}{!}{
\begin{tabular}{l|cccc}
\toprule
\multirow{2}{*}{Backbone} & \multirow{2}{*}{\begin{tabular}[c]{@{}c@{}}\#Params \\ (M)\end{tabular}} & \multirow{2}{*}{\begin{tabular}[c]{@{}c@{}}FLOPs\\ (G)\end{tabular}} & \multirow{2}{*}{\begin{tabular}[c]{@{}c@{}}IMNet\\ top-1 acc.\end{tabular}} & \multirow{2}{*}{\begin{tabular}[c]{@{}c@{}}City\\ mIoU\end{tabular}}\\\\ \midrule
HRNet18-MiT                  & 8.4          & 29.3    & 79.3&  80.30    \\
HRNet18-CSWin                &  8.1          & 22.3    & 79.5  & 80.95  \\
\textbf{\ours-b1}                   & \textbf{8.1}           & \textbf{14.1} & \textbf{80.5}  & \textbf{81.63}     \\ \midrule
HRNet32-MiT                  & 24.4          &  52.4   &  81.5  & 82.05 \\
HRNet32-CSWin                &  23.9          & 42.2    & 81.1  &  82.11 \\
\textbf{\ours-b2}                   & \textbf{20.8}          & \textbf{27.4}  & \textbf{82.3} & \textbf{82.81}     \\ \midrule
HRNet40-MiT                  & 40.1          &  108.0   & 82.3 & 82.10    \\
HRNet40-CSWin                &  39.5          & 96.3    & 82.4  &  82.38 \\
\textbf{\ours-b3}                   & \textbf{28.6}          & \textbf{66.8}  & \textbf{82.8} & \textbf{83.16}    \\
\bottomrule
Avg Improv.                   & -14.4\%          & -38.2\%  & +0.92 & +0.89    \\\bottomrule
\end{tabular}
}
\caption{Compare vanilla HRNet-ViT baselines with \ours on ImageNet-1K and Cityscapes \texttt{val}.
With heterogeneous branch designs and optimized blocks, \ours is more efficient than the vanilla HRNet-MiT and HRNet-CSWin.
}
\label{tab:CompareArch}
\end{table}

In Table~\ref{tab:CompareArch}, we directly replace residual blocks in HRNetV2 with MiT/CSWin Transformer blocks, which we refer to as a vanilla baseline.
When comparing HRNet-MiT with the sequential MiT, we notice the HR variants have comparable mIoUs while significantly saving hardware cost.
This shows that the \emph{multi-branch architecture is indeed helpful to boost the multi-scale representability}. 
However, the vanilla HRNet-ViT baseline overlooks the expensive cost of Transformers and is not efficient as the hardware cost quickly outweighs its performance gain.
In contrast, \ours benefits from \emph{heterogeneous branches and optimized components with less computation, fewer parameters, and enhanced model representability} than the vanilla HRNet-ViT baselines.

\noindent\textbf{Different window sizes.}~
\begin{table}[]
\centering
\resizebox{0.48\textwidth}{!}{%
\begin{tabular}{l|ccccc}
\toprule
window size $s$   & 7     & 9     & 11    & 13    & 15    \\ \midrule
GFLOPs         & 66.28 & 66.78 & 67.09 & 68.07 & 69.22 \\
Cityscapes mIoU (\%) & 82.82 & \textbf{83.16} & 83.15 & 82.88 & 82.90 \\ \bottomrule
\end{tabular}%
}
\caption{Evaluate \ours-b3 on Cityscapes \texttt{val} with different window sizes on the MR and LR paths.}
\label{tab:CompareWindow}
\end{table}

In Table~\ref{tab:CompareWindow}, we evaluate \ours-b3 on Cityscapes with different window sizes on the 3rd (MR) and 4th (LR) paths.
In general, different window sizes give similar mIoUs, while window sizes of 7 and 9 show the best performance-efficiency trade-off.
Increasing the window size from 7 to 9 helps \ours-b3 achieve +0.34 mIoU improvement with only 0.8\% more FLOPs.
However, overly-large window sizes bring no performance benefits with unnecessary computation overhead.
For example, further enlarging the window size from 9 to 15 causes 0.26 mIoU drop and 3.7\% more FLOPs.

\section{Related Work}
\label{sec:RelatedWork}
\noindent\textbf{Multi-scale representation learning for semantic segmentation.}~
Previous segmentation frameworks progressively down-sample the feature map to compute the LR representations~\cite{NN_CVPR2015_Long,NN_ECCV2018_Chen, NN_ICLR2021_Dosovitskiy}, and recover the HR features via up-sampling, e.g., SegNet~\cite{NN_TPAMI2017_Badrinarayanan}, UNet~\cite{NN_MICCAI2015_Ronneberger}, Hourglass~\cite{NN_ECCV2016_Newell}.
HRNet~\cite{NN_TPAMI2021_Wang} maintains the HR representations throughout the network with cross-resolution fusion. 
Lite-HRNet~\cite{NN_CVPR2021_Yu} proposes conditional channel weighting blocks to exchange information across resolutions.
HR-NAS~\cite{NN_CVPR2021_Ding} searches the channel/head settings for inverted residual blocks and the auxiliary Transformer branches.
HRFormer~\cite{NN_NeurIPS2021_Yuan} improves HRNetV2 by replacing residual blocks with Swin Transformer blocks.
Different from the convolutional HRNet-family, \ours is a pure ViT backbone with a novel \emph{multi-branch topology} that benefits both from HR architectures and self-attentions.
Distinguished from the direct CONV-to-Attention substitution in HRFormer, we explore a novel \emph{heterogeneous branch} design and various \emph{block optimization} techniques with higher performance and efficiency.

\noindent\textbf{Multi-scale ViT backbones.}~
Several multi-scale ViTs adopt hierarchical architectures to generate progressively down-sampled pyramid features~\cite{NN_ICCV2021_Wang, NN_ICCV2021_Chen,NN_NeurIPS2021_Chu,NN_ICCV2021_Fan,NN_NeurIPS2021_Xie,NN_Arxiv2021_Dong}.
For example, PVT~\cite{NN_ICCV2021_Wang} integrates a pyramid structure into ViTs for multi-scale feature extraction.
Twins~\cite{NN_NeurIPS2021_Chu} interleaves local and global attentions to learn multi-scale representations.
SegFormer~\cite{NN_NeurIPS2021_Xie} proposes an efficient hierarchical encoder to extract coarse and fine features.
CSWin~\cite{NN_Arxiv2021_Dong} further improves the performance with multi-scale cross-shaped local attentions.
However, they still follow the design concept of classification networks with a sequential topology. 
There is no information flow from LR to HR path inside those sequential ViTs, and the HR features are still very shallow ones of relatively low quality.
In contrast, our \ours adopts a multi-branch topology with enhanced multi-scale representability and improved efficiency.

\section{Conclusion}
\label{sec:Conclusion}
In this paper, we delve into the multi-scale representation learning in ViTs and present an efficient multi-scale high-resolution ViT backbone design, named \ours, for semantic segmentation.
We enhance ViTs with a multi-branch architecture to learn high-quality HR representations via cross-scale interaction.
To scale up \ours with high efficiency, we introduce heterogeneous branch designs and jointly optimize key building blocks with efficient embedding layers, augmented cross-shaped attentions, and mixed-scale convolutional FFNs.
In our evaluation, we observe that the multi-branch architecture can effectively boost the semantic segmentation performance of ViTs.
Besides, we find that branch-block co-optimization is the key to improving the efficiency of HR-ViT integration.
Experiments show that \ours outperforms SoTA ViT backbones on semantic segmentation with significant performance improvement and efficiency boost.
As a future direction, we look forward to evaluating \ours on more dense prediction vision tasks, e.g., object detection, to thoroughly demonstrate the potential of \ours as a strong vision backbone.

\clearpage
\appendix
\section{Detailed experimental settings}
\label{sec:AppendixExpSetting}
\subsection{Image classification on ImageNet-1K}
\label{sec:AppendixImageNet}
We pre-train \ours on ImageNet-1K for image classification.
To generate logits for classification, we append a classification head from HRNetV2 at the end of the \ours backbone.
Four multi-scale outputs from \ours are fed into convolutional bottleneck blocks, and the channels are mapped to 128, 256, 512, and 1024.
Then, we use stride-2 CONV3x3 to down-sample the highest resolution by 2$\times$ and double the channels by 2$\times$, and we add it to the next smaller resolution.
We repeat this process until we have the smallest resolution features.
Finally, we project the 1024-channel feature to 2048 channels via CONV1x1, followed by a global average pooling and a linear classifier.

We adopt a default image resolution of 224$\times$224 and train \ours with AdamW~\cite{NN_ICLR2019_Loshchilov} optimizer for 300 epochs using a cosine learning rate decay schedule, 20 epochs of linear warm-up, and an initial learning rate of 1e-3, a mini-batch size of 1,024, a weight decay rate of 0.05.
We set the weight decay rate for BatchNorm layers to 0.
We employ gradient clipping with a maximum magnitude of 1.
We employ label smoothing with a rate of 0.1 and various data augmentation methods used in DeiT~\cite{NN_ICML2021_Touvron}, including RandAugment~\cite{NN_CVPRW2020_Cubuk},
Cutmix~\cite{NN_ICCV2019_Yun},
Mixup~\cite{NN_Arxiv2017_Zhang}, and random erasing~\cite{NN_AAAI2020_Zhong}.
Note that model EMA and repeated augmentation are not employed here.
We use a stochastic drop path rate of 0.1 for all backbones.
For the classification head in \ours, we use a dropout rate of 0.1 for all variants.
Models are trained on 32 NVIDIA V100 GPUs with 32 images per GPU.

\subsection{Semantic segmentation on ADE20K and Cityscapes}
\label{sec:AppendixSegmentation}
ADE20K~\cite{NN_ADE20K} is a semantic segmentation dataset with 150 semantic categories.
It contains 25K images in total, including 20K images for training, 2K for validation, and the rest 3K for testing. 
Cityscapes~\cite{NN_Cityscapes} dataset contains 5000 fine-annotated high-resolution images with 19 categories.
The training image size for ADE20K and Cityscapes are cropped to 512$\times$512 and 1024$\times$1024, respectively.
For data augmentation, we use random horizontal flipping, random re-scaling within the ratio range of [0.5, 2.0], and random photometric distortion.
On ADE20K, the stochastic drop path rates are set to 0.1, 0.1, 0.25 for \ours-b1, \ours-b2, and \ours-b3, respectively.
On Cityscapes, the stochastic drop path rates are set to 0.15, 0.15, 0.25 for \ours-b1, \ours-b2, and \ours-b3, respectively.
We use an AdamW optimizer for 160 k iterations using a 'poly' learning rate schedule, 1,500 steps of linear warm-up, an initial learning rate of 6e-5, and a weight decay rate of 0.01.
The mini-batch size is set to 16 and 8 for ADE20K and Cityscapes, respectively.
We set the weight decay rate for BatchNorm layers to 0 and increase the initial learning rate for the segmentation head by 10$\times$.
Models on ADE20K are trained on 8 NVIDIA V100 GPUs with 2 images per GPU.
Models on Cityscapes are trained on 8 NVIDIA V100 GPUs with 1 image per GPU.
During inference, the image size for ADE20K \texttt{val} and Cityscapes \texttt{val} is set to 512$\times$2048 and 1024$\times$2048, respectively.
We do inference on Cityscapes with sliding window test by cropping 1024$\times$1024 patches.

\section{More experiments}
\label{sec:AppendixMoreExp}
\subsection{Semantic segmentation on ADE20K with UperNet head}
\label{sec:AppendixUperNet}
Besides the lightweight SegFormer head, UperNet~\cite{NN_ECCV2018_Xiao} is another segmentation framework that is widely used.
We evaluate \ours on ADE20K \texttt{val} with SegFormer and UperNet head in Table~\ref{tab:CompareADE20K}.
Our \ours can mostly maintain the advantages, but the computation benefits are not as significant as with the SegFormer head.
The reason is that the UperNet head dominates the computations ($>$89\%) and parameters ($>$54\%) in the cost breakdown. 
Hence any slimming on backbones will be considerably diluted. 
Moreover, we observe similar performance with lightweight SegFormer head and heavy UperNet head on \ours. 
Moreover, we do not observe more performance benefits from UperNet head than the SegFormer head on \ours.
One explanation is that \emph{\ours already has enough multi-resolution fusion, which makes the additional pyramid fusion in UperNet head less effective than used in the sequential architectures}. 
Hence, the lightweight SegFormer head is more suitable to \ours.

\subsection{Different block assignment strategies}
\label{sec:AppendixBlockAssign}
\begin{table}[hbp]
\centering
\resizebox{0.4\textwidth}{!}{
\begin{tabular}{c|cc}
\toprule
\begin{tabular}[c]{@{}c@{}}Block Assignment \\ on the 3rd Branch\end{tabular} & \begin{tabular}[c]{@{}c@{}}ImageNet-1K \\ Top-1 Acc\end{tabular} & \begin{tabular}[c]{@{}c@{}}Cityscapes \texttt{val} \\ mIoU\end{tabular} \\ \midrule
6-6-6-2                            & 80.53                 & 81.63               \\
8-8-2-2                            & 80.51                 & 81.50               \\
9-9-1-1                            & 80.50                 & 81.25               \\
17-1-1-1                           & 80.11                 & 81.27               \\ \bottomrule
\end{tabular}
}
\caption{Compare different block assignment strategies on the third LR branch in \ours-b1.}
\label{tab:AppendixCompareBlockAssign}
\end{table}

\begin{table*}[htp]
\centering
\resizebox{0.8\textwidth}{!}{
\begin{tabular}{l|ccc|ccc}
\toprule
                           & \multicolumn{3}{c|}{SegFormer Head~\cite{NN_NeurIPS2021_Xie}} & \multicolumn{3}{c}{UperNet Head~\cite{NN_ECCV2018_Xiao}}  \\
\multirow{-2}{*}{Backbone} & \#Param. (M)$\downarrow$  & GFLOPs$\downarrow$  & mIoU (\%)$\uparrow$ & \#Param. (M)$\downarrow$ & GFLOPs$\downarrow$ & mIoU (\%)$\uparrow$ \\ \midrule
MiT-B0~\cite{NN_NeurIPS2021_Xie}                     & 3.8           & 8.4     & 37.40     & -            & -      & -         \\
MiT-B1~\cite{NN_NeurIPS2021_Xie}                     & 13.7          & 15.9    & 42.20     & -            & -      & -         \\
CSWin-Ti~\cite{NN_Arxiv2021_Dong}                   & 5.9           & 11.4    & 41.43     & 33.7         & 216    & 44.41     \\
\textbf{\ours-b1}                   & \textbf{8.2}           & \textbf{14.6}    & \textbf{45.88}     & \textbf{35.9}         & \textbf{219}    & \textbf{47.19}     \\ \midrule
Twins-S~\cite{NN_NeurIPS2021_Chu}                    & -             & -       & -         & 54.4         & 219    & 46.20     \\
Swin-T~\cite{NN_ICCV2021_Liu}                     & -             & -       & -         & 59.9         & 234    & 44.50     \\
MiT-B2~\cite{NN_NeurIPS2021_Xie}                     & 27.5          & 62.4    & 46.50     & -            & -      & -         \\
CSWin-T~\cite{NN_Arxiv2021_Dong}                    & 22.4          & 28.3    & 47.88     & 59.9         & 234    & 49.30     \\
\textbf{\ours-b2}                   & \textbf{20.8}          & \textbf{28.0}    & \textbf{48.76}     & \textbf{49.7}         & \textbf{233}    & \textbf{49.10}     \\ \midrule
Twins-B~\cite{NN_NeurIPS2021_Chu}                    & -             & -       & -         & 88.5         & 248    & 47.70     \\
Swin-S~\cite{NN_ICCV2021_Liu}                    & -             & -       & -         & 81.3         & 253    & 47.60     \\
MiT-B3~\cite{NN_NeurIPS2021_Xie}                     & 47.3          & 79.0    & 49.40     & -            & -      & -         \\
CSWin-S~\cite{NN_Arxiv2021_Dong}                    & 37.3          & 78.1    & 49.93     & 64.6         & 247    & 50.00     \\
\textbf{\ours-b3}                   & \textbf{28.7}          & \textbf{67.9}    & \textbf{50.20}     & \textbf{55.4}         & \textbf{236}    & \textbf{50.04}     \\ \bottomrule
Avg improv.                   & -25.2\%          & -19.7\%    & +2.06     & -25.6\%         & -2.7\%    & +1.54     \\ \bottomrule
\end{tabular}
}
\caption{Performance and efficiency comparison of different ViT backbones on the ADE20K \texttt{val} segmentation dataset.
Average improvements of \ours over baselines are summarized for each framework. 
}
\label{tab:CompareADE20K}
\end{table*}

In Table~\ref{tab:AppendixCompareBlockAssign}, we compare different block assignment strategies on the 3rd low-resolution path in \ours-b1 on ImageNet-1K and Cityscapes.
We observe a clear trend that when concentrating more blocks in one module, e.g., from 6-6-6-2 to 17-1-1-1, the benefits from the cross-resolution fusion in the HR architecture diminish accordingly, leading to degraded ImageNet classification accuracy and Cityscapes segmentation mIoU.
This phenomenon can be attributed to the information loss in the deep LR module, which validates the effectiveness of our \emph{even block assignment} strategy with better information interaction and detail preservation.

\end{document}